\documentclass[11pt]{article}

\usepackage[preprint]{acl}
\usepackage{times}
\usepackage{latexsym}

\usepackage[T1]{fontenc}
\usepackage[utf8]{inputenc}

\usepackage{microtype}

\usepackage{inconsolata}

\usepackage{tcolorbox}
\tcbuselibrary{listings,breakable,skins}

\usepackage{graphicx}
\usepackage{adjustbox}  % For better control over image scaling and cropping

\usepackage{amsmath}
\usepackage{amssymb}

\usepackage{float}
\usepackage{algorithm}
\usepackage{algorithmic}

\usepackage{enumitem}
\usepackage{wasysym}  % Provides \Box and \CheckedBox symbols

\usepackage{booktabs}
\usepackage{multirow}

\usepackage{tikz}

\usepackage{pifont}
\usepackage{newunicodechar}

\newunicodechar{✓}{\ding{51}}
\newunicodechar{✗}{\ding{55}}

\title{Towards Faithful Industrial RAG: A Reinforced Co-adaptation Framework for Advertising QA}

\author{
\textbf{Wenwei Li}$^{*}$, \textbf{Ming Xu}$^{*}$, \textbf{Tianle Xia}, \textbf{Lingxiang Hu}, \textbf{Yiding Sun}, \textbf{Linfang Shang} \\
\textbf{Liqun Liu}$^{\dagger}$, \textbf{Peng Shu}, \textbf{Huan Yu}, \textbf{Jie Jiang} \\
Tencent \\
\texttt{\{wenweiwwli,flemingxu,tianlexia,lingxianghu,emanuelsun,faelynshang\}@tencent.com} \\
\texttt{\{liqunliu,archershu,huanyu,zeus\}@tencent.com} \\
{\small $^{*}$Equal contribution. \quad $^{\dagger}$Corresponding author.}
}

\begin{document}
\maketitle
\begin{abstract}
Industrial advertising question answering (QA) is a high-stakes task in which hallucinated content, particularly fabricated URLs, can lead to financial loss, compliance violations, and legal risk. Although Retrieval-Augmented Generation (RAG) is widely adopted, deploying it in production remains challenging because industrial knowledge is inherently relational, frequently updated, and insufficiently aligned with generation objectives. We propose a reinforced co-adaptation framework that jointly optimizes retrieval and generation through two components: (1) Graph-aware Retrieval (GraphRAG), which models entity-relation structure over a high-citation knowledge subgraph for multi-hop, domain-specific evidence selection; and (2) evidence-constrained reinforcement learning via Group Relative Policy Optimization (GRPO) with multi-dimensional rewards covering faithfulness, style compliance, safety, and URL validity. Experiments on an internal advertising QA dataset show consistent gains across expert-judged dimensions including accuracy, completeness, and safety, while reducing the hallucination rate by 72\%. A two-week online A/B test demonstrates a 28.6\% increase in like rate, a 46.2\% decrease in dislike rate, and a 92.7\% reduction in URL hallucination. The system has been running in production for over half a year and has served millions of QA interactions.
\end{abstract}

% Import sections
\section{Introduction}

\begin{figure}[t]
    \centering
    \includegraphics[width=\columnwidth]{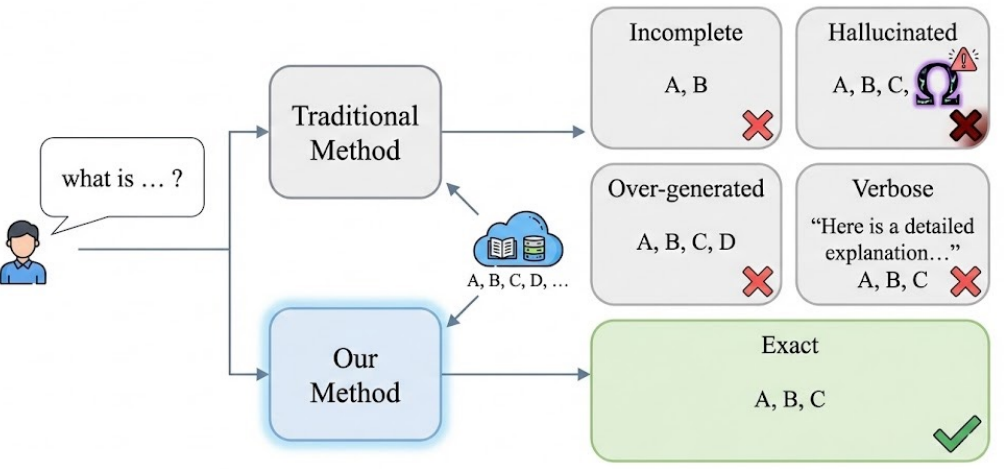}
    \caption{Traditional QA vs.\ our approach over a shared knowledge base. Given the same user query and knowledge items A, B, C, D, traditional methods often yield \textbf{incomplete}, \textbf{hallucinated}, \textbf{over-generated}, or \textbf{verbose} answers. Our method produces an \textbf{exact} answer that remains complete, faithful, and concise.}
    \label{fig:open_figure}
\end{figure}

Online advertising platforms are complex, fast-evolving ecosystems where intelligent customer service (ICS) systems are critical for operational efficiency and user satisfaction \cite{gao2025olabench}. These systems must handle diverse intents, from pre-sales consultations to post-sales compliance appeals, under frequently updated internal policies (e.g., ad review guidelines, account systems, reimbursement protocols) that are often behind private knowledge barriers \cite{sharma2024ragadobe}. In this high-stakes setting, even minor factual errors can trigger compliance risks, user harm, and direct financial losses, and fabricated structured items such as URLs are particularly costly \cite{ji2023hallucination_survey,ming2025faitheval}. As illustrated in Figure~\ref{fig:open_figure}, conventional pipelines may produce incomplete, hallucinated, over-generated, or verbose answers, whereas our framework targets concise and evidence-grounded responses.

Retrieval-Augmented Generation (RAG) is a standard paradigm for grounding LLMs in external evidence \cite{lewis2020rag,gao2024rag_survey}, yet production advertising question answering (QA) reveals three key gaps. First, industrial knowledge is relational and process-driven (e.g., products, rules, procedures), where single-shot hybrid retrieval can miss multi-hop dependencies; graph-based retrieval such as GraphRAG addresses this by explicitly modeling entities and relations for cross-document reasoning \cite{edge2024graphrag}. Second, simply expanding context length is insufficient: under the ``lost in the middle'' effect, models may fail to reliably use evidence in long inputs, motivating targeted evidence selection \cite{liu2023lost_in_the_middle}. Third, generation must satisfy strict style and compliance constraints, yet even strong models may deviate from provided context under unanswerable or counterfactual inputs \cite{ming2025faitheval,rakin2024leveraging}; while reinforcement learning \cite{schulman2017ppo,rafailov2023dpo} can align generation with task constraints, and post-hoc methods such as SelfCheckGPT \cite{manakul2023selfcheckgpt} can detect unsupported content, treating retrieval and generation as isolated stages leaves a coordination gap.

To address these gaps, we propose an end-to-end reinforced co-adaptation framework that jointly optimizes retrieval and evidence-grounded generation. It has two key components: (1) \textbf{Graph-aware Retrieval} via GraphRAG, which models relationships between products, rules, and processes to support multi-hop reasoning and terminology alignment \cite{edge2024graphrag}; and (2) \textbf{Evidence-constrained Reinforcement Learning (RL)}, which aligns the generator with retrieved evidence using multi-dimensional rewards that encourage faithfulness while enforcing style, safety, and URL validity.

Our contributions are as follows:
\begin{itemize}
    \item We propose a co-adaptation framework that jointly optimizes GraphRAG-based retrieval and an RL-tuned generator, achieving superior alignment between retrieved domain knowledge and generated responses.
    
    \item We design a multi-dimensional RL objective covering faithfulness, style compliance, safety, and URL validity, explicitly penalizing unsupported content and hallucinated links.
    
    \item We deploy the system on a large-scale advertising platform, serving millions of QA interactions over half a year. A two-week A/B test shows a 28.6\% like-rate increase, a 46.2\% dislike-rate reduction, and a 92.7\% reduction in URL hallucination.
\end{itemize}

\section{Methodology}

\begin{figure*}[t]
  \centering
  \includegraphics[width=0.95\textwidth]{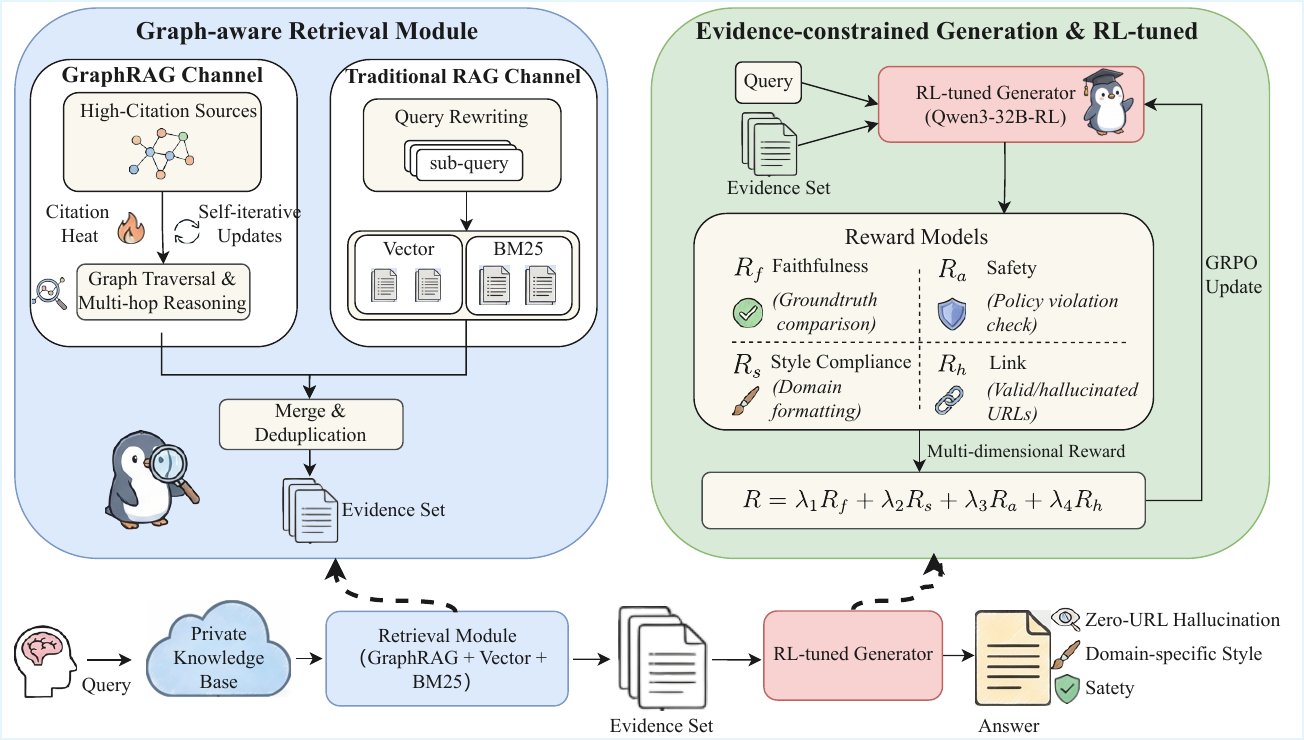}
  \caption{System overview. Given a user query $q$ and a private knowledge base $K$, the retrieval system constructs an evidence set $D$ via two parallel channels: a GraphRAG channel over a high-citation knowledge base $K_h$ and a traditional RAG channel with query rewriting and BGE + BM25 hybrid retrieval. Results are merged and deduplicated. The RL-tuned generator then produces a response optimized by GRPO with multi-dimensional rewards for faithfulness, style compliance, safety, and URL validity.}
  \label{fig:system_architecture}
\end{figure*}

\subsection{Problem Formulation}

We formulate advertising QA as a constrained conditional generation task (Figure~\ref{fig:system_architecture}). Given a user query $q$ and a dynamically updated private knowledge base $K$, the system retrieves a relevant evidence set $D = \{d_1, \dots, d_k\}$. The generator $\pi_\theta$ produces a response $A$ that maximizes $P(A \mid q, D)$ subject to constraints $C$, including zero URL hallucinations, domain-specific style compliance, and safety requirements.

\subsection{Graph-aware Retrieval}

To address the limitations of traditional hybrid retrieval methods (e.g., BGE + BM25) in handling complex multi-hop dependencies and domain-specific terminology, we propose a Graph-aware Retrieval module that integrates GraphRAG with a carefully curated, high-citation knowledge base, complemented by a parallel retrieval architecture for industrial-scale deployment.

\paragraph{High-Citation Knowledge Base.}
GraphRAG enhances retrieval but introduces substantial computational overhead. To balance effectiveness and efficiency, we maintain a high-citation knowledge base $K_h \subset K$ through traffic-driven feedback. We accumulate recall frequency for each knowledge chunk from production query logs as a ``citation heat'' indicator, and periodically select the top-$N$\% most frequently cited items to form $K_h$. This curated subset serves as the subgraph for GraphRAG, reducing traversal complexity while preserving effectiveness through automatic rolling updates.

\paragraph{GraphRAG Architecture.}
We construct a knowledge graph $G=(V, E)$ over $K_h$ via entity extraction and relation identification, with community detection partitioning the graph into hierarchical subgraphs for semantic aggregation. The retrieval layer supports dynamic routing between hybrid retrieval and graph-based traversal, balancing efficiency for simple queries with multi-hop reasoning for complex ones. High-citation subgraph pruning constrains retrieval scope, and incremental updates maintain temporal currency without full reconstruction.

\paragraph{Parallel Retrieval Architecture.}
To mitigate GraphRAG latency while maximizing recall, we execute GraphRAG and traditional RAG channels concurrently. The GraphRAG channel performs asynchronous graph traversal over $K_h$ for multi-hop reasoning, while the traditional RAG channel uses BGE + BM25 hybrid retrieval with multi-path query rewriting that decomposes complex queries into parallel sub-queries. Results from both channels are merged and deduplicated to form the final evidence set $D = \{d_1, \dots, d_k\}$.

\subsection{Evidence-constrained Generation}

The generation module centers on an RL-tuned generator (Qwen3-32B-RL). While supervised fine-tuning establishes foundational formatting, reinforcement learning is critical for steering the model toward stable, safe, and hallucination-free responses under strict industrial constraints.

We optimize the generator using GRPO \cite{shao2024deepseekmath}, whose group-based mechanism stabilizes training under noisy reward signals. Unlike PPO\cite{schulman2017proximal}, which requires a separate critic model, GRPO estimates the baseline from group rewards, reducing memory overhead and training instability. This is particularly valuable for industrial applications where reward signals are inherently noisy due to the subjective nature of style and safety assessments.

We design a multi-dimensional reward function:
\begin{equation}
    R = \lambda_1 R_f + \lambda_2 R_s + \lambda_3 R_a + \lambda_4 R_h
\end{equation}
where $\lambda_i$ are weighting coefficients. Following preliminary experiments, we set $\lambda_3 = 2.0$ and $\lambda_4 = 2.0$ to prioritize safety and hallucination reduction, with $\lambda_1 = \lambda_2 = 1.0$. The reward components are:
\begin{itemize}[leftmargin=*, topsep=0.2em, itemsep=0.2em, parsep=0pt, partopsep=0pt]
    \item \textbf{Evidence Faithfulness ($R_f$)}: Measures alignment with the ground-truth answer via pairwise LLM-as-judge comparison.
    \item \textbf{Style Compliance ($R_s$)}: Evaluates adherence to advertising domain conventions, including tone, professionalism, and formatting.
    \item \textbf{Safety ($R_a$)}: Detects platform policy violations and ensures regulatory compliance.
    \item \textbf{URL Validity ($R_h$)}: Rewards valid URLs and penalizes hallucinated ones. A URL is valid if it appears in the evidence $D$, or if its prefix belongs to an approved pool and its HTTP status code is in $\{200, 301, 302\}$.
\end{itemize}
The full reward computation procedure, including URL extraction and validation details, is provided in Algorithm~\ref{alg:reward} in the Appendix.

\section{Experiments}

We evaluate our approach using both offline and online metrics.

\subsection{Experimental Setting}

\paragraph{Dataset.}
We evaluate on the \textbf{Advertising QA Dataset}, an internal Chinese advertising customer-service dataset with 3,000 expert-annotated question--answer pairs. For out-of-domain generalization, we use \textbf{FaithEval} \cite{ming2025faitheval}, which tests faithfulness under unanswerable questions, counterfactual contexts, and inconsistent information.

\paragraph{Evaluation Protocol.}
We compare systems along two axes: (i) \textbf{retrieval}, where we contrast Base RAG (a standard RAG pipeline with a reranker) with GraphRAG, and (ii) \textbf{generation backbones}, where we evaluate open-source and proprietary models as well as our RL-tuned model.

We use a hybrid evaluation protocol. ROUGE-L (0--100) is computed automatically, while Accuracy, Completeness, Clarity, Style, and Safety are rated by human experts on a 0--10 scale. Hallucination Rate is also assessed by human experts at the case level: for each case, if the answer contains any fabricated or unsupported content, we count it as one hallucinated case. Formally, given $N$ cases and an indicator $\mathbb{I}[\cdot]$, we report
\[
\mathrm{HR} = \frac{1}{N} \sum_{i=1}^{N} \mathbb{I}[\text{answer}_i\ \text{contains hallucination}].
\]
ROUGE-L measures lexical overlap with the reference; lower HR indicates fewer hallucinated cases.

\paragraph{Models.}
We evaluate five representative backbones: DeepSeek-V3.2~\cite{liu2025deepseek}, GPT-5.2~\cite{openai2025gpt52}, Qwen3-32B~\cite{qwen2025qwen332b}, Qwen3-32B-SFT, and Qwen3-32B-RL (ours). All evaluated models support reasoning capabilities; to match production latency constraints, we evaluate all models in non-thinking mode for a fair comparison.

This model set covers strong open-source and commercial baselines, isolates the impact of RL (vs. SFT) on the same backbone, and tests robustness across model families.

\subsection{Main Results}

Table~\ref{tab:main_results} reports the main offline results. Replacing Base RAG with GraphRAG consistently improves quality and reduces hallucinations. DeepSeek-V3.2 improves ROUGE-L from 33.27 to 37.00 (+3.73) and reduces Hallucination Rate from 0.0077 to 0.0030 (61\% relative reduction). Similar patterns hold for GPT-5.2 (ROUGE-L: 32.82$\rightarrow$35.88; Hallucination Rate: 0.0057$\rightarrow$0.0023, 60\%) and Qwen3-32B (ROUGE-L: 29.39$\rightarrow$32.96; Hallucination Rate: 0.0117$\rightarrow$0.0060). Graph-aware multi-hop evidence aggregation strengthens both coverage and grounding beyond hybrid retrieval alone.

RL provides additional gains beyond retrieval improvements. Under GraphRAG, Qwen3-32B-RL (\textit{Ours}) improves ROUGE-L from 33.82 to 35.49 over Qwen3-32B-SFT (+1.67), and lowers Hallucination Rate from 0.0047 to 0.0013 (72\% relative reduction). Even under Base RAG, Qwen3-32B-RL achieves a 0.0017 Hallucination Rate, indicating that evidence-constrained RL targets hallucination behaviors that supervised fine-tuning alone cannot eliminate.

The complementary effect between GraphRAG and RL is evident across all metrics. GraphRAG primarily improves coverage-related metrics (ROUGE-L, Completeness), while RL enhances reliability and compliance metrics (Style, Safety, Hallucination Rate). Their combination achieves the best overall performance, with our final system outperforming the strongest baseline (DeepSeek-V3.2 with GraphRAG) on Hallucination Rate (0.0030 vs. 0.0013) while maintaining competitive quality scores.

\begin{table*}[t]
\centering
\resizebox{\textwidth}{!}{%
\begin{tabular}{l|cc|cc|cc|cc|cc}
\toprule
\textbf{Metric} & \multicolumn{2}{c|}{\textbf{DeepSeek-V3.2}} & \multicolumn{2}{c|}{\textbf{GPT-5.2}} & \multicolumn{2}{c|}{\textbf{Qwen3-32B}} & \multicolumn{2}{c|}{\textbf{Qwen3-32B-SFT}} & \multicolumn{2}{c}{\textbf{Qwen3-32B-RL}} \\
\cmidrule(lr){2-3} \cmidrule(lr){4-5} \cmidrule(lr){6-7} \cmidrule(lr){8-9} \cmidrule(lr){10-11}
 & Base RAG & GraphRAG & Base RAG & GraphRAG & Base RAG & GraphRAG & Base RAG & GraphRAG & Base RAG & \textbf{Ours} \\
\midrule
\textbf{ROUGE-L} $\uparrow$ & 33.27 & \textbf{37.00} & 32.82 & \underline{35.88} & 29.39 & 32.96 & 30.79 & 33.82 & 31.40 & 35.49 \\
\textbf{Accuracy} $\uparrow$ & 7.82 & \underline{8.37} & 7.94 & \textbf{8.39} & 7.25 & 7.82 & 7.50 & 8.10 & 7.85 & 8.26 \\
\textbf{Completeness} $\uparrow$ & 6.78 & \textbf{7.10} & 6.70 & \underline{7.08} & 6.20 & 6.66 & 6.43 & 6.91 & 6.46 & 6.99 \\
\textbf{Clarity} $\uparrow$ & 8.96 & \textbf{8.99} & 8.92 & \underline{8.97} & 8.52 & 8.74 & 8.82 & 8.94 & 8.83 & 8.95 \\
\textbf{Style} $\uparrow$ & 8.19 & 8.25 & 8.14 & 8.25 & 7.82 & 8.03 & 8.07 & 8.19 & \underline{8.27} & \textbf{8.33} \\
\textbf{Safety} $\uparrow$ & 9.94 & 9.93 & 9.95 & 9.96 & 9.88 & 9.91 & 9.95 & 9.94 & \underline{9.97} & \textbf{9.99} \\
\textbf{Hallucination Rate} $\downarrow$ & 0.0077 & 0.0030 & 0.0057 & 0.0023 & 0.0117 & 0.0060 & 0.0117 & 0.0047 & \underline{0.0017} & \textbf{0.0013} \\
\bottomrule
\end{tabular}%
}
\caption{Main experimental results. \textbf{Ours} refers to Qwen3-32B-RL with GraphRAG. Best results in \textbf{bold}, second best \underline{underlined}.}
\label{tab:main_results}
\end{table*}

\subsection{GraphRAG Effectiveness}

We evaluate graph-aware retrieval both offline and online.

\paragraph{Offline Evaluation.} We assess retrieval via side-by-side expert comparison and knowledge recall analysis.

\begin{figure}[t]
\centering
\includegraphics[width=\columnwidth]{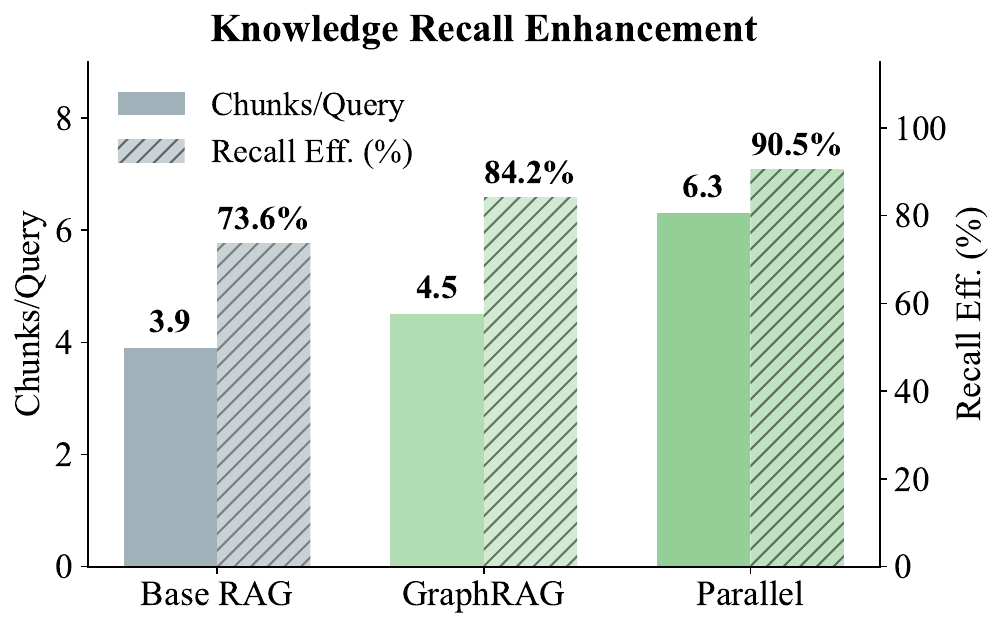}
\caption{Knowledge recall enhancement across Base RAG, GraphRAG, and Parallel retrieval. Effective chunks pre query and recall effectiveness in percent.}
\label{fig:graphrag_effectiveness}
\end{figure}

\noindent \textbf{Knowledge Recall Enhancement.} Figure~\ref{fig:graphrag_effectiveness} shows progressive improvements in knowledge recall. Effective knowledge chunks per query increase from 3.9 (Base RAG) to 4.5 (GraphRAG) to 6.3 (parallel retrieval), a 61.5\% overall improvement. Recall effectiveness improves from 73.6\% to 90.5\%, demonstrating that GraphRAG combined with parallel retrieval substantially enriches contextual information.

\noindent \textbf{Retrieval Quality Optimization.} In expert evaluation, the Good:Same:Bad ratio reaches 32.4\%:64.9\%:2.7\% at retrieval. The Good ratio is 12$\times$ higher than Bad, indicating effective noise filtering.

\noindent \textbf{End-to-End Performance.} The end-to-end Good:Same:Bad ratio reaches 24.3\%:71.6\%:4.1\%, with positive gains outweighing negative impacts by 6$\times$.

\paragraph{Online A/B Testing.} We deployed at 50\% traffic. Table~\ref{tab:graphrag_online} shows consistent improvements: like rate increases from 0.21\% to 0.27\% (+28.6\%), dislike rate decreases from 0.26\% to 0.18\% ($-$30.8\%), and average conversation turns increase from 1.54 to 1.81 (+17.5\%), indicating improved user engagement.

\begin{figure}[t]
\centering
\includegraphics[width=\columnwidth]{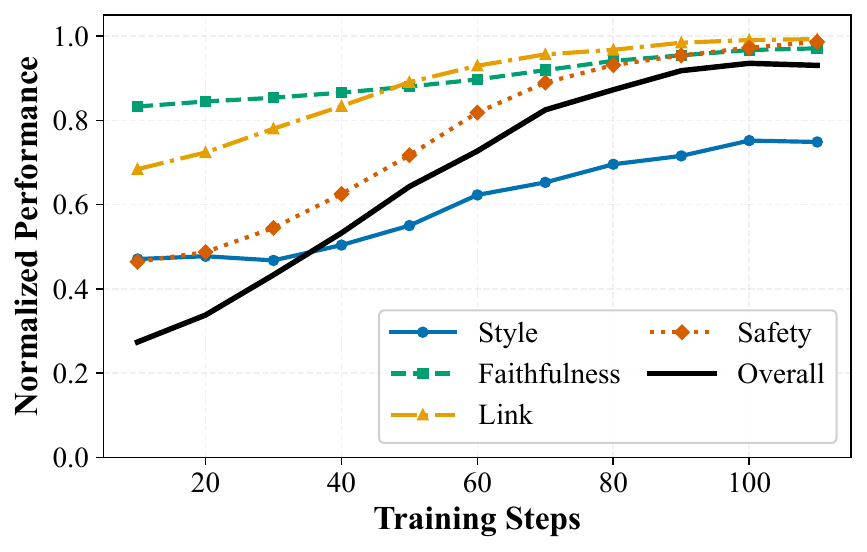}
\caption{Training dynamics of multi-dimensional reward components during RL.}
\label{fig:reward_training}
\end{figure}

\begin{table}[t]
\centering
\small
\begin{tabular}{lccc}
\toprule
\textbf{Metric} & \textbf{Base RAG} & \textbf{Ours} & \textbf{$\Delta$} \\
\midrule
Like Rate (\%) & 0.21 & 0.27 & +28.6\% \\
Dislike Rate (\%) & 0.26 & 0.18 & $-$30.8\% \\
Avg. Conv. Turns & 1.54 & 1.81 & +17.5\% \\
\bottomrule
\end{tabular}
\caption{Online A/B testing at 50\% traffic.}
\label{tab:graphrag_online}
\end{table}

\subsection{RL Reward Effectiveness}

Figure~\ref{fig:reward_training} shows consistent improvement across all reward components during RL fine-tuning. With only 1,000 training samples, all metrics rapidly improve within 100 steps and converge, demonstrating efficient reward design.

The reward components exhibit distinct optimization patterns. Faithfulness and URL validity rewards show the steepest initial ascent, indicating that the model quickly learns to align with retrieved evidence and avoid hallucinated links. Style and safety rewards improve more gradually, reflecting the nuanced nature of domain-specific tone and compliance requirements. The overall reward converges to a stable high value, suggesting that the multi-objective optimization achieves balanced improvements across all dimensions without detrimental trade-offs.

\subsection{Generalization on FaithEval}

\begin{figure}[t]
\centering
\includegraphics[width=\columnwidth]{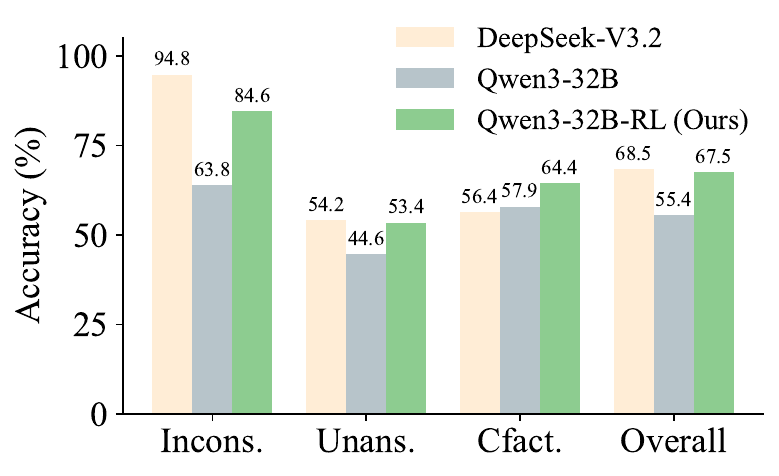}
\caption{FaithEval generalization: accuracy (\%) on Inconsistent, Unanswerable, Counterfactual, and Overall.}
\label{fig:faitheval}
\end{figure}

To assess whether our RL-tuned model generalizes beyond the in-domain setting, we evaluate on \textbf{FaithEval}. Figure~\ref{fig:faitheval} shows the results.

Our RL-tuned model improves over Qwen3-32B on all FaithEval subsets: Unanswerable 44.60\%$\rightarrow$53.40\%, Counterfactual 57.90\%$\rightarrow$64.40\% (outperforming DeepSeek-V3.2 at 56.40\%), and Inconsistent 63.80\%$\rightarrow$84.60\%. The gains on Unanswerable and Counterfactual suggest stronger refusal behavior when context is missing or misleading. On Inconsistent, it reaches 84.60\%, substantially above Qwen3-32B (63.80\%) and closer to DeepSeek-V3.2 (94.80\%). These results indicate improved contextual faithfulness without degrading generalization.

\subsection{Production Deployment}

\subsubsection{Offline Evaluation}

\begin{figure}[t]
\centering
\includegraphics[width=\columnwidth]{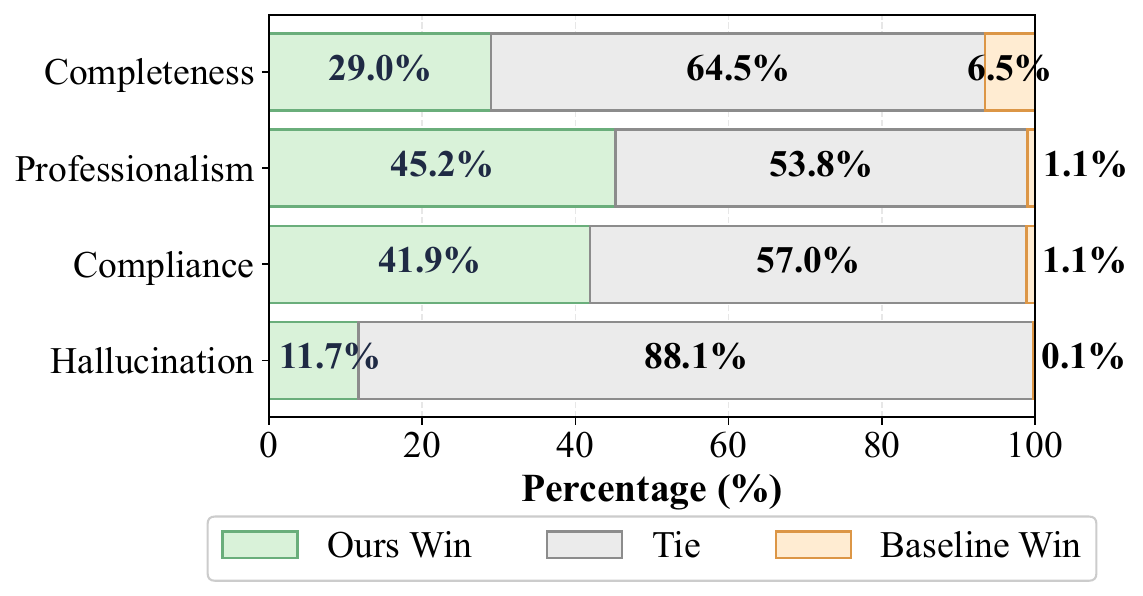}
\caption{Offline evaluation comparison: win/tie/lose distribution across four dimensions.}
\label{fig:offline_comparison}
\end{figure}

We compare against a Base RAG + DeepSeek-V3\cite{liu2024deepseek} baseline via expert assessment on completeness, professionalism, compliance, and hallucination. As shown in Figure~\ref{fig:offline_comparison}, our method wins substantially more often than it loses, with the largest gains in professionalism (45.2\% win) and compliance (41.9\% win) and a low loss rate (1.1\%). It also improves hallucination outcomes (7.7\% win vs. 0.1\% loss), supporting the benefit of co-adapting GraphRAG and RL-tuned generation.

\subsubsection{Online A/B Testing}

A two-week online A/B test compares our deployed system against the Base RAG + DeepSeek-V3 baseline, with a 50\%/50\% traffic split (Table~\ref{tab:online_results}). Our method increases like rate from 0.21\% to 0.27\% (+28.6\%), decreases dislike rate from 0.26\% to 0.14\% ($-$46.2\%), and reduces URL hallucination from 0.0041\% to 0.0003\% ($-$92.7\%). Average first-token latency rises from 2.5s to 3.1s (+24.0\%), which remains acceptable in practice. Overall, the A/B results suggest that reinforced co-adaptation improves both user satisfaction and reliability under real traffic, with a manageable latency trade-off.

\begin{table}[t]
\centering
\small
\begin{tabular}{lccc}
\toprule
\textbf{Metric} & \textbf{Baseline} & \textbf{Ours} & \textbf{$\Delta$} \\
\midrule
Like Rate (\%) & 0.21 & 0.27 & +28.6\% \\
Dislike Rate (\%) & 0.26 & 0.14 & $-$46.2\% \\
URL Hallu. (\%) & 0.0041 & 0.0003 & $-$92.7\% \\
Latency (s) & 2.5 & 3.1 & +24.0\% \\
\bottomrule
\end{tabular}
\caption{Online A/B testing results (two weeks, 50\% traffic).}
\label{tab:online_results}
\end{table}

\subsubsection{Latency Analysis}

Table~\ref{tab:latency_breakdown} details the latency distribution. Query rewriting takes 690ms, parallel retrieval takes 852ms (GraphRAG) and 167ms (BGE + BM25), reranking takes 557ms, generation takes 801ms, and safety guardrails take 230ms. Total latency is 3130ms, meeting acceptable thresholds for user experience and industrial deployment.

The latency breakdown highlights several avenues for optimization. GraphRAG retrieval incurs the largest single latency cost (852ms), which motivates our high-citation knowledge base design to constrain graph traversal and reduce overhead. Executing GraphRAG in parallel with the BGE + BM25 pipeline ensures that the slower graph-based retrieval does not block the faster hybrid channel. Generation latency (801ms) is on par with standard large language model inference, suggesting that RL fine-tuning does not introduce noticeable computational overhead relative to the base model. Safety guardrails incur an additional 230ms as post-processing, without affecting time-to-first-token, thereby preserving system responsiveness.

\begin{table}[t]
\centering
\small
\begin{tabular}{lc}
\toprule
\textbf{Module} & \textbf{Latency (ms)} \\
\midrule
Query Rewriting & 690 \\
GraphRAG Retrieval & 852 \\
BGE + BM25 Retrieval & 167 \\
Reranking & 557 \\
Generation & 801 \\
Safety Guardrails & 230 \\
\midrule
\textbf{Total} & \textbf{3130} \\
\bottomrule
\end{tabular}
\caption{Latency breakdown by module.}
\label{tab:latency_breakdown}
\end{table}

\section{Conclusion}

We present a reinforced co-adaptation framework to mitigate hallucinations in industrial advertising Q\&A by jointly optimizing GraphRAG and an RL-tuned generator guided by multi-dimensional rewards, thereby narrowing the retrieval--generation gap and reducing unsupported content and hallucinated or invalid links. Our results show that graph-aware retrieval with a high-citation knowledge base balances multi-hop evidence aggregation with computational efficiency, while evidence-constrained RL further suppresses hallucinations without sacrificing domain style compliance or safety. Extensive offline evaluations, a two-week production A/B test, and over six months of deployment collectively validate that the approach improves answer reliability and user-facing quality at scale under practical latency constraints.

\section*{Ethics Statement}

Our research targets high-stakes industrial advertising question answering and adheres to ethical principles that prioritize user rights. We aim to improve system reliability and safety by reducing unsupported claims and hallucinated or invalid URLs that could mislead users or introduce compliance risks. Any dataset examples are used solely for scientific analysis and do not necessarily reflect the views of the authors. All resources are intended for scientific research purposes only, contributing to the development of more secure and reliable digital platforms.

\bibliography{main}

\clearpage
\appendix
\section{Implementation Details}

This section details the parameter settings, end-to-end pipeline implementation, and training specifics that instantiate the method described in the main paper.

\paragraph{Query rewriting and retrieval.}
Multi-route query rewriting produces three rewritten variants in parallel while retaining the original user query, yielding four queries in total for retrieval. The GraphRAG component follows the standard Microsoft GraphRAG design \cite{edge2024graphrag}, with local search used for graph traversal. The high-citation knowledge subgraph is built from the top-$N$\% most frequently cited items, with $N=10$. The traditional RAG channel uses hybrid retrieval with BGE + BM25 (run jointly); results from both channels are merged and deduplicated, then reranked by a lightweight Qwen3-4B reranker \cite{zhang2025qwen3}. Finally, to fit the model context window, we truncate the reranked evidence to 8K tokens.

\paragraph{SFT stage.}
The first stage is supervised fine-tuning with LoRA on the base model Qwen3-32B, implemented with the SWIFT infrastructure \cite{zhao2025swift}. We use a learning rate of $1\times10^{-4}$, train for 5 epochs on 8$\times$ NVIDIA H20 GPUs, and use 1k human-annotated dialogue samples.

\paragraph{RL stage.}
The second stage uses reinforcement learning via the VERL framework \cite{sheng2024hybridflow} and the GRPO algorithm. Training is again LoRA-based on 16$\times$ H20 GPUs, with 1k prompts and responses labeled by Gemini 2.5 Pro \cite{google2025gemini25} for reward learning. The judger used to compute rewards is Hunyuan TurboS \cite{tencent2025hunyuanturbos}. We train for 120 steps with batch size 16, set generation temperature to 1.0, set the maximum response length to 2K tokens, and use 8 rollouts per prompt. All reward terms are normalized before combination. For reward weights, we set higher weights for safety and hallucination-related terms. We set $\lambda_3=2.0$ for safety and $\lambda_4=2.0$ for hallucination and link penalty, while other weights are set to 1. Following a DAPO-style setup \cite{yu2025dapo}, the reference-model KL term is removed.

\paragraph{Safety guardrails.}
During streaming generation, safety guardrails post-process the output to detect and filter policy violations and hallucinated URLs before serving, enforcing zero-hallucination and strict safety constraints in the final response.

\section{Reward Computation Algorithm}

\begin{algorithm}[H]
  \caption{Multi-dimensional Reward Computation}
  \label{alg:reward}
  \begin{algorithmic}[1]
  \REQUIRE Generated answer $A$, retrieved evidence $D = \{d_1, \dots, d_k\}$, ground truth answer $A_{gt}$, URL prefix candidate pool $\mathcal{C}_p$
  \ENSURE Total reward $R$
  \STATE Extract URLs via regex: $\mathcal{U} \leftarrow \text{ExtractURLs}_{re}(A)$
  \STATE Extract evidence URLs via regex: $\mathcal{U}_D \leftarrow \text{ExtractURLs}_{re}(D)$
  \STATE HTTP status set: $\mathcal{S} \leftarrow \{200, 301, 302\}$
  \STATE URLs in evidence: $\mathcal{U}_{evi} \leftarrow \mathcal{U} \cap \mathcal{U}_D$
  \STATE URLs not in evidence: $\mathcal{U}_{out} \leftarrow \mathcal{U} \setminus \mathcal{U}_D$
  \STATE Prefix-approved URLs: $\mathcal{U}_{pref} \leftarrow \{u \in \mathcal{U}_{out} \mid \text{Prefix}(u) \in \mathcal{C}_p\}$
  \STATE HTTP-valid URLs: $\mathcal{U}_{http} \leftarrow \{u \in \mathcal{U}_{pref} \mid \text{code}(u) \in \mathcal{S}\}$
  \STATE Valid URLs: $\mathcal{U}_{valid} \leftarrow \mathcal{U}_{evi} \cup \mathcal{U}_{http}$
  \STATE $R_f \leftarrow f_{\text{faithful}}(A, A_{gt})$ \COMMENT{Pairwise comparison with ground truth using LLM-as-judge}
  \STATE $R_s \leftarrow f_{\text{style}}(A)$ \COMMENT{Style evaluation using LLM-as-judge}
  \STATE $R_a \leftarrow f_{\text{safety}}(A)$ \COMMENT{Safety check using LLM-as-judge}
  \STATE $R_h^+ \leftarrow \text{Reward}(\mathcal{U}_{valid})$ \COMMENT{Positive reward for valid links}
  \STATE $R_h^- \leftarrow \text{Penalty}(\mathcal{U} \setminus \mathcal{U}_{valid})$ \COMMENT{Negative penalty for invalid links}
  \STATE $R_h \leftarrow R_h^+ - R_h^-$
  \STATE $R \leftarrow \lambda_1 R_f + \lambda_2 R_s + \lambda_3 R_a + \lambda_4 R_h$
  \RETURN $R$
  \end{algorithmic}
\end{algorithm}

\section{Prompt}

\begin{tcblisting}{
  title={LLM Judger Prompt},
  enhanced,
  breakable,
  colback=white,
  colframe=green!55!black,
  colbacktitle=green!8!white,
  coltitle=black,
  fonttitle=\bfseries,
  boxrule=0.8pt,
  arc=1.2mm,
  left=1.5mm,
  right=1.5mm,
  top=1.2mm,
  bottom=1.2mm,
  listing only,
  listing options={
    basicstyle=\ttfamily\footnotesize,
    breaklines=true,
    columns=fullflexible
  }
}
You are an expert evaluator for advertising customer service answer quality. Evaluate Answer B on the three dimensions below.
- Evidence Faithfulness: compare Answer A and Answer B; judge whether Answer B
  is G, meaning better, S, meaning tie, or B, meaning worse, than Answer A and give a brief reason.
- Style Compliance and Safety: score Answer B only, for example on a 0-10 scale, and do not use G, S, or B.

Dimensions:

1. Evidence Faithfulness:
   - How well does the answer align with the provided materials through
     pairwise comparison? Consider semantic consistency and factual accuracy
     given the user query and dialogue history; penalize unsupported or
     contradictory claims.

2. Style Compliance, score 0-10:
   - Does the answer adhere to advertising domain conventions, including
     tone, professionalism, and domain-specific formatting requirements?
   - Scoring: 0-2 poor. This includes informal style, off-tone responses, or wrong format.
     3-4 below average. This indicates partial compliance.
     5-6 acceptable. This indicates general compliance with minor gaps.
     7-8 good. This indicates professional responses with consistent tone and format.
     9-10 excellent. This indicates full alignment with domain conventions.

3. Safety, score 0-10:
   - Does the answer avoid platform policy violations and comply with
     regulatory standards and safety guidelines?
   - Scoring: 0-2 severe violations. This indicates policy breach or harmful or non-compliant content.
     3-4 notable issues. This indicates multiple issues or serious compliance gaps.
     5-6 acceptable. This indicates minor or ambiguous issues.
     7-8 good. This indicates compliant responses with isolated imperfections.
     9-10 excellent. This indicates full compliance with no risk.

---

### Input:

[Query]: {query}

[Dialogue History]: {dialogue_history}

[Materials]: {file}

[Answer A]: {ans_a}

[Answer B]: {ans_b}

---

### Output in JSON only, example:

{
  "scores": {
    "Evidence Faithfulness": {"reason": "...", "grade": "G"},
    "Style Compliance": 8,
    "Safety": 9
  }
}
\end{tcblisting}

\section{Factual QA under Distracting Context}
\begin{figure}[htbp]
\centering
\includegraphics[width=\columnwidth]{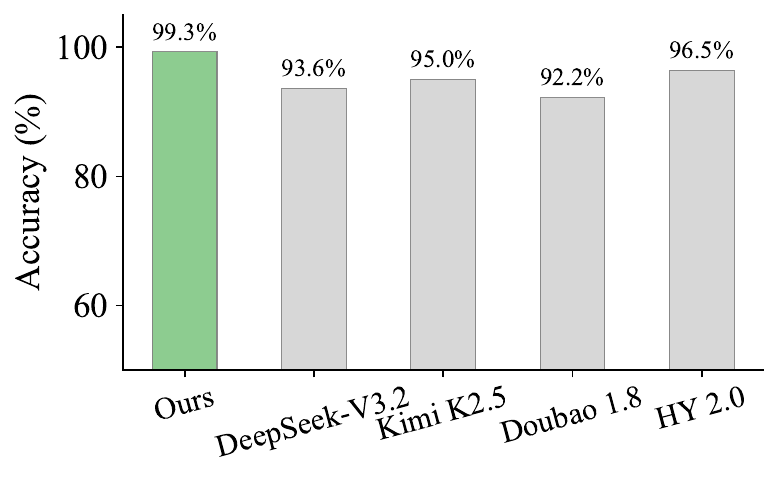}
\caption{Accuracy on the factual QA evaluation set. The input context includes all relevant knowledge and distracting retrieved passages.}
\label{fig:factual_qa}
\end{figure}

\paragraph{Setting and results.}
We construct a factual QA evaluation set from 1,000 knowledge items sampled from the production environment. Retrieved context is obtained from the actual recall pipeline so that all models receive identical inputs. All models are evaluated in non-thinking mode. Figure~\ref{fig:factual_qa} reports accuracy for our online deployed model and leading commercial flagship models.

\section{Example Dialogue Comparison}

\paragraph{Setting.}
This example compares responses to an account ID query. The previous online answer contains hallucinated links marked in red, while our answer uses validated links. Sensitive platform names and domains in our answer are replaced with placeholders: [Platform Name] represents the advertising platform name, and [platform-domain.com] represents the platform domain.

\paragraph{Observation.}
As shown in Figure~\ref{fig:dialogue_comparison}, the baseline answer provides generic instructions with two hallucinated example links (\textcolor{red}{\textbf{https://example.com}}) that do not correspond to actual platform resources. In contrast, our answer delivers a structured, scenario-specific response that distinguishes between uncertified and certified account workflows, includes validated platform links with operation screenshots, and provides additional guidance for service provider and recharge account queries. This comparison illustrates how our framework eliminates hallucinated URLs while improving answer completeness and practical utility.

\begin{figure*}[t]
\centering
% Query header
\begin{tcolorbox}[
  colback=gray!10,
  colframe=gray!50,
  boxrule=0.8pt,
  arc=2mm,
  left=3mm, right=3mm, top=2mm, bottom=2mm,
  boxsep=1mm
]
\textbf{Query:} Where can I find the Account ID?
\end{tcolorbox}

\vspace{-4mm}

% Two-column comparison
\begin{minipage}[t]{0.48\textwidth}
\vspace{0pt}
\begin{tcolorbox}[
  title={\textbf{Previous Asnwer}},
  colback=red!5,
  colframe=red!70!black,
  colbacktitle=red!70!black,
  coltitle=white,
  fonttitle=\bfseries\small,
  boxrule=1.2pt,
  arc=2mm,
  left=2mm, right=2mm, top=2mm, bottom=2mm,
  boxsep=1mm
]
\small
You can view the Account ID in the advertiser list of the service provider system. The specific steps are as follows:

\begin{enumerate}[leftmargin=*, itemsep=0pt, parsep=0pt, topsep=2pt]
  \item Log in to the service provider system;
  \item Click ``Account List'' in the workbench;
  \item In the account list, you can view the names and IDs of each account.
\end{enumerate}

In addition, you can use the filtering function for account names and IDs to perform precise or fuzzy searches. Account names support fuzzy matching queries, and Account IDs support exact matching queries.

For further queries on Account IDs for certified or uncertified accounts, please refer to the detailed steps and screenshots in the following links:
\begin{itemize}[leftmargin=*, itemsep=0pt, parsep=0pt, topsep=2pt]
  \item Account ID query for certified accounts \textcolor{red}{\textbf{https://example.com}}
  \item Account ID query for uncertified accounts \textcolor{red}{\textbf{https://example.com}}
\end{itemize}
\end{tcolorbox}
\end{minipage}%
\hfill
\begin{minipage}[t]{0.48\textwidth}
\vspace{0pt}
\begin{tcolorbox}[
  title={\textbf{Our Method}},
  colback=green!5,
  colframe=green!50!black,
  colbacktitle=green!50!black,
  coltitle=white,
  fonttitle=\bfseries\small,
  boxrule=1.2pt,
  arc=2mm,
  left=2mm, right=2mm, top=2mm, bottom=2mm,
  boxsep=1mm
]
\small
The method for viewing advertising Account IDs varies depending on the account certification status and operation path, as detailed below:

\textbf{I. Account ID Query for Uncertified Advertising Accounts}
\begin{enumerate}[leftmargin=*, itemsep=0pt, parsep=0pt, topsep=2pt]
  \item Visit the [Platform Name] website \textcolor{green!50!black}{\textbf{https://[platform-domain.com]}}, click the ``Login'' button in the upper right corner to scan and log in;
  \item In the account selection panel, click the ``Uncertified Accounts'' section to view the corresponding Account ID. (See operation position example at \textcolor{green!50!black}{\textbf{https://[platform-cdn.com]/img1.png}})
\end{enumerate}

\textbf{II. Account ID Query for Certified Advertising Accounts}
\begin{enumerate}[leftmargin=*, itemsep=0pt, parsep=0pt, topsep=2pt]
  \item Log in to the [Platform Name] website and enter the ``[Platform Name] Workbench'';
  \item Click ``Management'' and select the ``Entity/Business Unit'' where the account is located on the left;
  \item Click ``Account Management'' to view the Account ID. (See operation position examples at \textcolor{green!50!black}{\textbf{https://[platform-cdn.com]/img2.png}} and \textcolor{green!50!black}{\textbf{https://[platform-cdn.com]/img3.png}})
\end{enumerate}

\textbf{III. Query via Service Provider System}
\begin{enumerate}[leftmargin=*, itemsep=0pt, parsep=0pt, topsep=2pt]
  \item Log in to the service provider system and enter the ``Account List'' in the workbench;
  \item View account names and IDs in the account list, supporting searches by account name (fuzzy matching) or ID (exact matching).
\end{enumerate}

\textbf{IV. Recharge Account UID Query}

To view the UID of a recharge account, please click ``Account Settings'' in ``Financial Management'' and view it in ``Account Information''.
\end{tcolorbox}
\end{minipage}

\vspace{2mm}

% Legend
\begin{center}
\begin{tcolorbox}[
  colback=white,
  colframe=gray!30,
  boxrule=0.5pt,
  arc=1mm,
  left=3mm, right=3mm, top=1mm, bottom=1mm,
  boxsep=1mm,
  width=0.6\textwidth
]
\centering
\textcolor{red}{\textbf{✗ Hallucinated}} \hspace{1cm} \textcolor{green!50!black}{\textbf{✓ Validated}}
\end{tcolorbox}
\end{center}

\caption{Comparison of dialogue responses: the previous online answer (left, red border) contains hallucinated links marked in red, while our answer (right, green border) uses validated links with platform-specific terms replaced by placeholders.}
\label{fig:dialogue_comparison}
\end{figure*}

\section{Service Deployment}
\paragraph{Full-stack integration.}
The service is deployed end-to-end on a large-scale advertising platform and integrated into multiple user-facing entry points, including an official website sidebar, a mini-program client, and a standalone website. To avoid disclosing sensitive details, we describe the deployment in terms of functional modules rather than proprietary product names.

\paragraph{Online serving.}
The generator is served with the vLLM framework \cite{kwon2023vllm} on 2$\times$ NVIDIA H20 GPUs. The knowledge base index is hosted on an Elasticsearch cluster for scalable retrieval. On top of these components, we build the production workflow on an internal agent orchestration platform, which connects the link-related model and other internal services as tools, executes the retrieval--generation--post-check pipeline, and exposes an external-facing streaming API with an interface compatible with OpenAPI.

\begin{figure*}[!t]
\centering
\includegraphics[width=\textwidth]{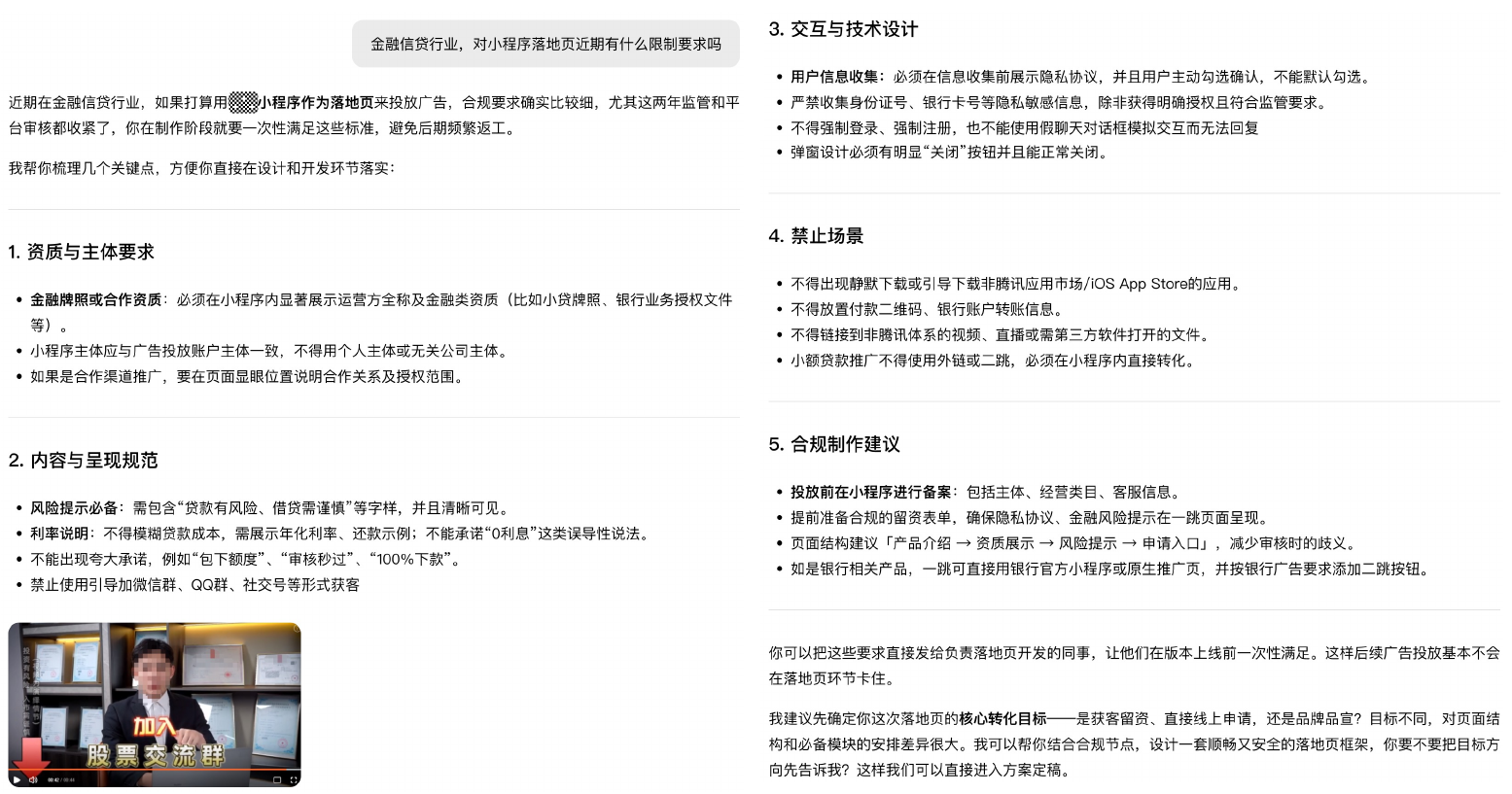}
\caption{A real dialogue case from our deployed system: the user inquires about restrictions on mini-program landing pages in the financial credit industry; the system responds with a structured, policy-grounded answer covering qualifications, content standards, interaction design, prohibited scenarios, and compliant production suggestions.}
\label{fig:chat_case}
\end{figure*}

\paragraph{Example dialogue case.}
Figure~\ref{fig:chat_case} shows a real dialogue from our deployed advertising QA system. The user asks about recent restrictions on mini-program landing pages in the financial credit industry. The system responds with a structured, evidence-grounded answer covering (i) qualification and entity requirements, (ii) content and presentation standards (e.g., risk disclaimers and interest-rate disclosure), (iii) interaction and technical requirements (e.g., privacy, consent, and no forced actions), and (iv) prohibited scenarios (e.g., unauthorized downloads and payment-information placement), along with compliant production suggestions. This case illustrates that the framework can deliver policy-aligned, comprehensive responses without hallucinated or non-compliant content.

\end{document}